\documentclass[conference]{IEEEtran}
\IEEEoverridecommandlockouts

\usepackage{cite}
\usepackage{amsmath,amssymb,amsfonts}
\usepackage{algorithmic}
\usepackage{graphicx}
\usepackage{textcomp}
\usepackage{xcolor}

\usepackage[ruled,vlined]{algorithm2e}  

\usepackage{amsthm}

\usepackage{graphicx}
\usepackage{booktabs}
\usepackage{multirow}
\usepackage{colortbl}  
\usepackage{rotating}  
\usepackage{pgfplots}  
\def\BibTeX{{\rm B\kern-.05em{\sc i\kern-.025em b}\kern-.08em
    T\kern-.1667em\lower.7ex\hbox{E}\kern-.125emX}}
\pgfplotsset{compat=1.18}  
\begin{document}

\title{Guided Prompt Evolution for Vision-Language Models Adaptation\\}

\author{Enming Zhang, Jiayang Li, Yanlong Wang, Yanru Wu, Zhenyu Liu, and Yang Li
\thanks{Enming Zhang, Jiayang Li, Yanru Wu, Zhenyu Liu are with Tsinghua Shenzhen International Graduate School, Tsinghua University. Yanlong Wang is with Sun Yat-sen University.
Yang Li is with Chinese University of Hong Kong, Shenzhen.
Corresponding author: Yang Li (email: yangl@cuhk.edu.cn).}}


\maketitle
\begin{abstract}
The adaptation of large-scale vision-language models (VLMs) to downstream tasks with limited labeled data remains a significant challenge. While parameter-efficient prompt learning methods offer a promising path, they often suffer from catastrophic forgetting of pre-trained knowledge. Toward addressing this limitation, our work is grounded in the insight that governing the evolutionary path of prompts is essential for forgetting-free adaptation. To this end, we propose EvoPrompt, a novel framework designed to explicitly steer the prompt trajectory for knowledge-preserving fine-tuning. Specifically, our approach employs a Modality-Shared Prompt Projector (MPP) to generate hierarchical prompts from a unified embedding space. Critically, an evolutionary training strategy decouples low-rank updates into directional and magnitude components, preserving early-learned semantic directions while only adapting their magnitude, thus enabling prompts to evolve without discarding foundational knowledge. Extensive experiments demonstrate that EvoPrompt achieves state-of-the-art performance in few-shot learning while robustly preserving the original zero-shot capabilities of pre-trained VLMs.
\end{abstract}


\section{Introduction}
\label{sec:intro}
Large-scale pre-trained vision-language models (VLMs) \cite{radford2021learning, jia2021scaling, yu2022coca, huang2023language, li2023blip2}, exemplified by works like CLIP \cite{radford2021learning} and ALIGN \cite{jia2021scaling}, have revolutionized zero-shot generalization across diverse downstream tasks, including image classification \cite{dosovitskiy2020image, fei2004learning}, visual question answering \cite{antol2015vqa, goyal2017making}, and cross-modal retrieval \cite{lee2018stacked, chen2020uniter}. Their success stems from learning highly transferable visual and linguistic representations through contrastive pre-training on massive web-scale datasets.

However, adapting these powerful models to specific downstream tasks with limited labeled samples presents a significant challenge. The conventional approach of full fine-tuning, which updates all model parameters, is often prohibitively expensive in terms of computation and storage, given the massive scale of VLMs \cite{alayrac2022flamingo, chen2024scaling}. To address this, parameter-efficient adaptation methods\cite{lester2021power}, particularly prompt learning, have gained prominence. Techniques like CoOp \cite{zhou2022learning} and CoCoOp \cite{zhou2022conditional} introduce a set of learnable continuous prompts while keeping the pre-trained backbone frozen, drastically reducing tunable parameters.

\begin{figure}[t!]
    \centering
    \includegraphics[width=\columnwidth]{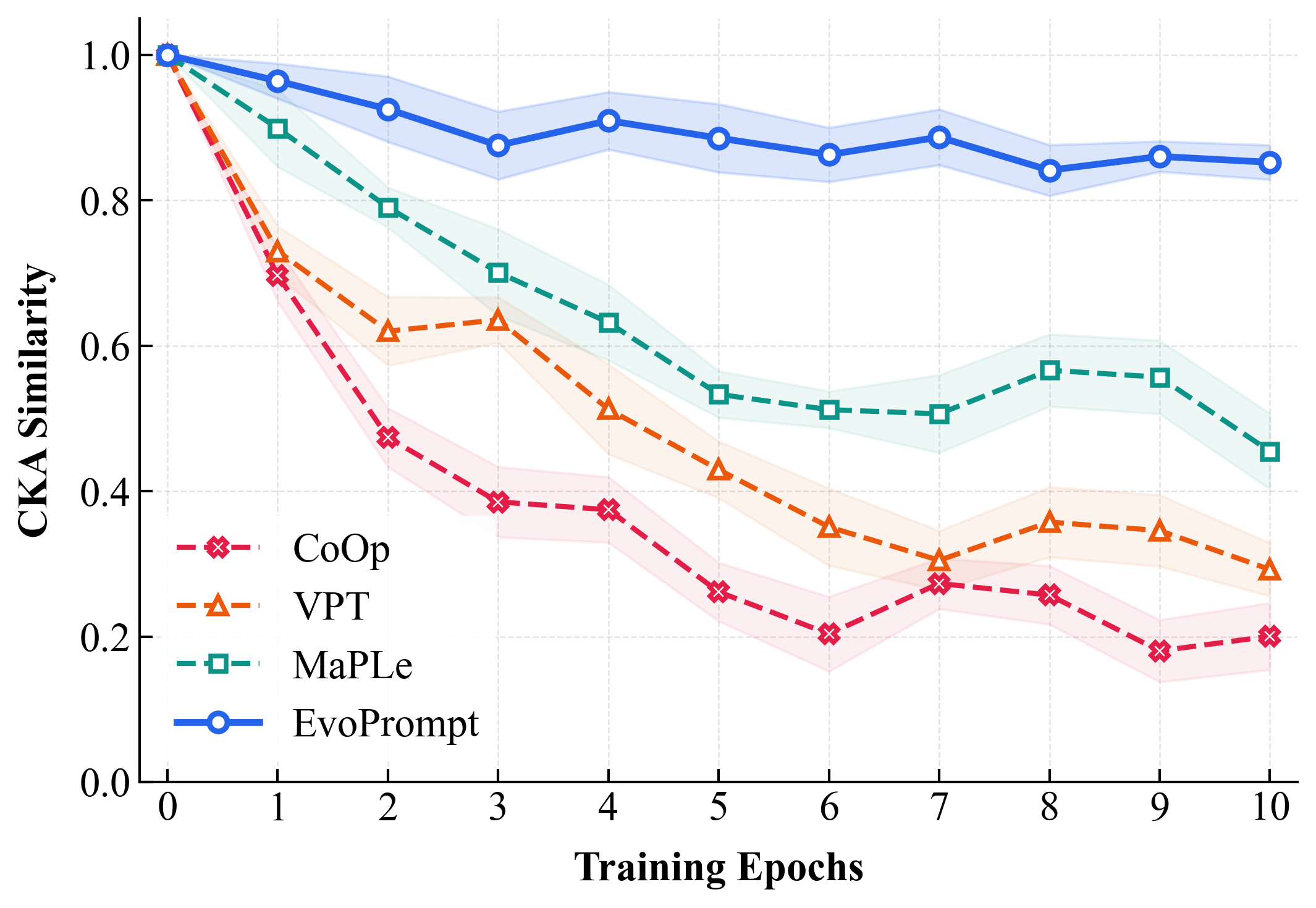}
    \caption{CKA similarity between prompt-tuned features at each 
training epoch and the frozen pre-trained VLM (Epoch~0). 
Baseline methods exhibit increasing representational 
drift from the pre-trained semantic space, whereas EvoPrompt 
maintains consistently structural similarity throughout training.}
    \label{fig:cka}
\end{figure}

\begin{figure*}[t!]
    \centering
    \includegraphics[alt={figure}, width=\textwidth]{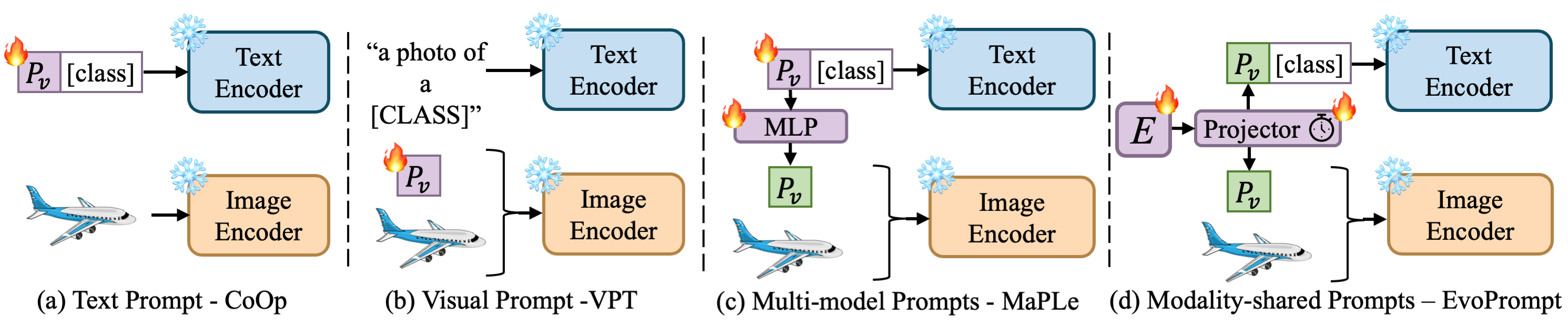}
    \caption{Comparison of our proposed EvoPrompt frameworks with related representative efficient transfer learning for VLMs.}
    \label{fig:first}
\end{figure*}

Despite this efficiency, a fundamental limitation has been largely 
overlooked: existing prompt tuning methods optimize prompts solely to minimize task loss, with no regard for their evolution during training. As shown in Fig.~\ref{fig:cka}, our Centered Kernel Alignment (CKA) analysis reveals that this unconstrained optimization induces progressive representational drift away from the pre-trained semantic space. Prompts progressively abandon their broad semantic structure, collapsing into narrow, task-specific representations—a manifestation of catastrophic forgetting~\cite{khattak2023self, wang2022dualprompt} that undermines the zero-shot generalization capabilities VLMs acquire through large-scale pre-training.

We therefore argue that governing the evolutionary trajectory of prompts, rather than merely optimizing their final representations, is the cornerstone of knowledge-preserving adaptation. As illustrated in Fig.~\ref{fig:first}, current frameworks directly optimize raw prompt parameters with no structural constraint, making their trajectories inherently unstable. In contrast, we propose EvoPrompt, which generates prompts through a shared embedding space to impose structural coherence over the entire training trajectory. Specifically, we introduce a Modality-Shared Prompt Projector (MPP) that replaces 
isolated per-layer prompts with a unified embedding projected into 
layer-specific representations, establishing structural coherence for 
stable evolution across layers and modalities. We further regulate 
the temporal dynamics of adaptation via a trajectory-aware training 
strategy that disentangles low-rank updates into directional and 
magnitude components. Semantic directions captured in early training 
are frozen, while only their magnitudes are refined, enabling 
task-specific specialization without discarding pre-trained knowledge. 
To further anchor the adapted features to the pre-trained semantic space, we introduce a lightweight knowledge constancy loss that constrains feature drift throughout training. Our contributions are summarized as follows:

\begin{itemize}
    \item We uncover the representational drift inherent in unconstrained prompt optimization, establishing trajectory governance as a cornerstone for robust VLM adaptation.

    \item We propose EvoPrompt, a trajectory-governed adaptation paradigm whose core is a trajectory-aware training strategy that factorizes prompt evolution into frozen semantic directions and adaptive magnitudes. 

    \item Extensive experiments on few-shot learning, cross-dataset transfer, 
    and domain generalization demonstrate state-of-the-art performance with 
    robust preservation of zero-shot generalization.
\end{itemize}

We will release the code and pretrained models upon acceptance of the paper.

\section{Related Work}
\subsection{Vision-Language Models}
The landscape of computer vision has been profoundly reshaped by the advent of Vision-Language Models (VLMs), which forge robust semantic connections between visual and textual data. A myriad of foundational architectures---such as CLIP \cite{radford2021learning}, ALIGN \cite{jia2021scaling}, BLIP-2 \cite{li2023blip2}, and Flamingo \cite{alayrac2022flamingo}---have been trained on web-scale, paired multi-modal datasets using self-supervised contrastive objectives \cite{oord2018representation}. By assimilating unprecedented volumes of training samples \cite{schuhmann2022laion}, these foundation models learn highly transferable feature representations, yielding impressive zero-shot inference capabilities across a broad array of applications. Despite these triumphs, effectively transferring such massive architectures to specialized target distributions with limited annotated data remains remarkably challenging. To navigate this data-scarce regime, a vast corpus of literature has focused on tailoring pre-trained VLMs to diverse downstream scenarios, encompassing few-shot image classification \cite{zhang2022tip}, object detection \cite{du2022learning}, and semantic segmentation \cite{rao2022denseclip, dong2023zegclip}.
\subsection{Efficient Transfer Learning for VLMs}
Parameter-efficient fine-tuning (PEFT), particularly via prompting techniques, originated in the natural language domain to steer massive linguistic models without updating their full parameter space \cite{houlsby2019parameter, li2021prefix, liu-etal-2022-p, hu2021lora}. This philosophy was subsequently extended to multi-modal learning \cite{gao2024clip, zhang2022tip, chen2022adaptformer, sung2022vl}, enabling the rapid adaptation of frozen VLMs. Pioneering works like CoOp \cite{zhou2022learning} appended optimizable continuous tokens to the textual branch of CLIP, while CoCoOp \cite{zhou2022conditional} conditioned these tokens on visual inputs to mitigate overfitting to seen classes. To further regulate the optimization process and retain foundational knowledge, methods such as KgCoOp \cite{yao2023visual} penalize the discrepancy between the learned textual embeddings and the original frozen embeddings, and PLOT \cite{chen2023plot} leverages optimal transport to holistically match vision and text semantics. Moving beyond unimodal prompting, MaPLe \cite{khattak2023maple} synchronizes the adaptation by explicitly linking deep learnable tokens across both vision and text encoders. More recently, approaches like PromptSRC \cite{khattak2023self} and TCP \cite{yao2024tcp} have integrated self-consistency mechanisms and class-aware regularization to further constrain the learning trajectory. While these prompt-driven methodologies have proven highly effective in adapting large-scale models with minimal overhead, they suffer from inherent structural limitations.

\section{Method: EvoPrompt}
We develop our EvoPrompt framework on top of the pre-trained CLIP \cite{radford2021learning}. In the following, we first introduce preliminary knowledge on CLIP and then present our proposed EvoPrompt.

\subsection{Preliminaries}
We provide a brief overview of the CLIP model's notation and core operations used in our method. CLIP contains two primary components: a visual encoder $F$ and a text encoder $G$.

\textbf{Image Feature Extraction.} The visual encoder $f$ is structured as a Vision Transformer (ViT) with $L$ consecutive transformer blocks, $\{F_i\}_{i=1}^{L}$. An input RGB image $I \in \mathbb{R}^{H \times W \times 3}$ is split into $M$ non-overlapping patches. A linear projection layer maps each patch to a $d_v$-dimensional vector, producing the initial patch embedding matrix $E_0 \in \mathbb{R}^{M \times d_v}$. This matrix is prepended with a learnable [CLS] token $c_0$, combined with positional encodings, and fed into the transformer stack. The processing at the $i$-th layer is formulated as:
\begin{equation}
    [c_i, E_i] = F_i([c_{i-1}, E_{i-1}]), \quad i=1, \dots, L.
\end{equation}
The final [CLS] token representation $c_L$ is projected via a linear layer $\Phi_v$ to obtain the image feature vector $f^v = \Phi_v(c_L) \in \mathbb{R}^d$.

\textbf{Text Feature Extraction.} For a text prompt $T$ (e.g., “a photo of a [CLASS]"), the input is first tokenized into a sequence of $N$ tokens. These tokens are embedded as $T_0 \in \mathbb{R}^{N \times d_t}$ and concatenated with special [SOS] and [EOS] tokens $b_0$ and $e_0$, along with positional encodings. The sequence is processed by $L$ transformer layers $\{ G_i \}_{i=1}^{L}$ in the text encoder:
\begin{equation}
    [b_i, T_i, e_i] = G_i([b_{i-1}, T_{i-1}, e_{i-1}]), \quad i=1, \dots, L.
\end{equation}
The final [EOS] token representation $e_L$ is linearly projected to obtain the text feature vector $f^t = \Phi_t(e_L) \in \mathbb{R}^d$.

\textbf{Zero-Shot Classification and Optimization.} For a $C$-class task, we construct $C$ text prompts to obtain text features $\{f^t_c\}_{c=1}^C$. Given an image feature $f^v$, the prediction probability for class $c$ is computed via cosine similarity and a softmax with temperature $\tau$:

\begin{equation}
    p(y=c \mid f^v) = \frac{\exp(s_c / \tau)}{\sum_{j=1}^{C} \exp(s_j / \tau)}, \quad  \textit{where}  \quad s_c = \frac{{f^v}^\top f^t_c}{\|f^v\| \|f^t_c\|},
\end{equation}
where $\top$ denotes the vector transpose operation. The model is typically optimized using the cross-entropy loss. For a training sample with ground-truth label $y$, the loss is defined as:

\begin{equation}
    \mathcal{L}_{InfoNCE}(f^v, f^t) = -\log \frac{\exp(s_y / \tau)}{\sum_{j=1}^{C} \exp(s_j / \tau)}.
\end{equation}

\subsection{Modality-Shared Prompt Projector} 
\label{sec:MPP}

Previous multimodal prompting schemes, such as MaPLe~\cite{khattak2023maple}, typically insert prompts into each layer independently. While this provides layer-specific guidance, such isolated prompts often prevent the model from distilling and propagating beneficial information across the hierarchical depth of the encoders. We argue that prompts should capture the hierarchical semantic progression across consecutive layers and maintain a degree of inter-layer correlation. Furthermore, leveraging complementary information across modalities can enrich the prompt generation process. To this end, we propose the Modality-Shared Prompt Projector (MPP), which jointly fosters cross-layer information flow and complementary cross-modal interaction.

\begin{figure*}[t!]
    \centering
    \includegraphics[width=\textwidth]{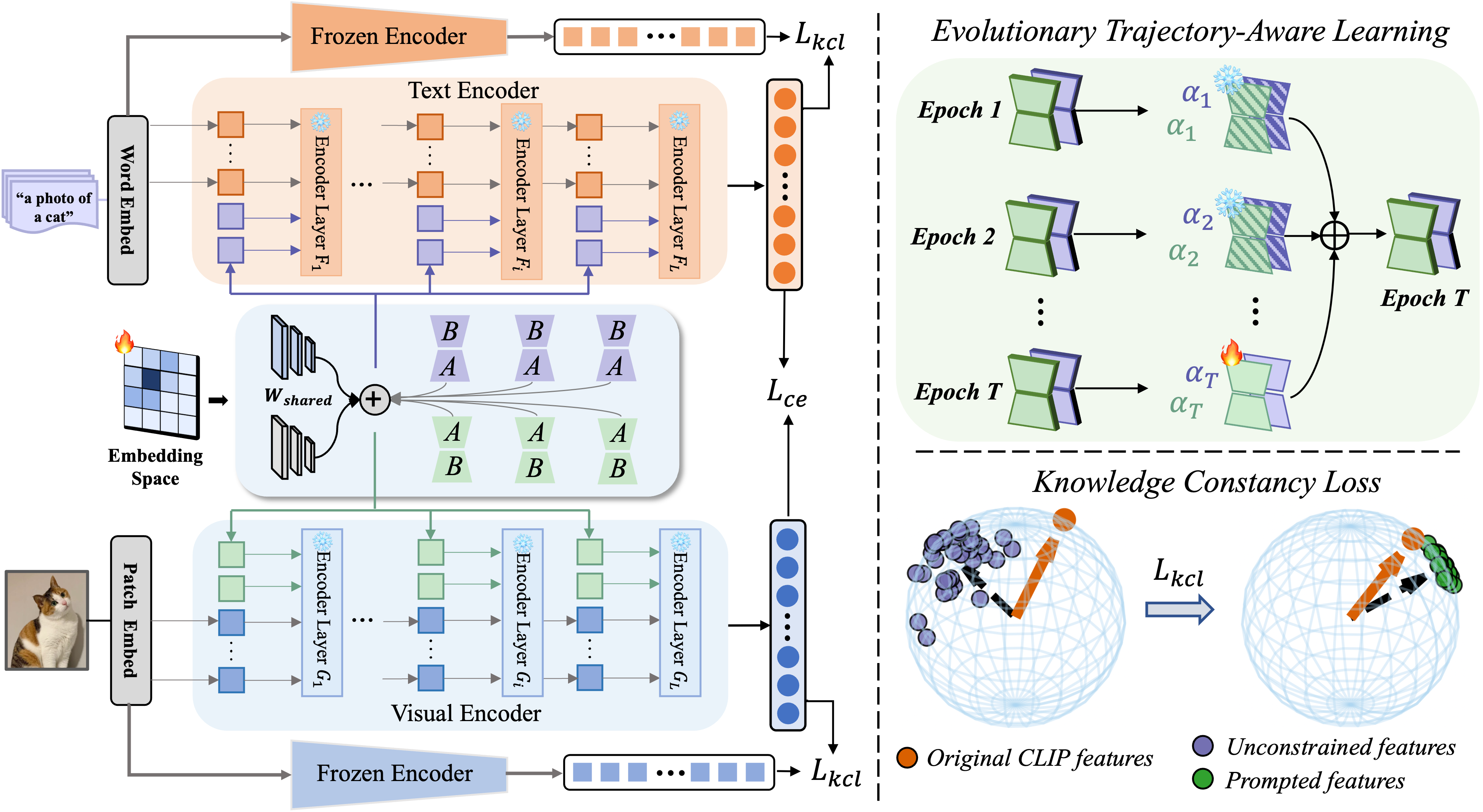}
    \caption{Overview of the proposed EvoPrompt framework. \textbf{Left}: Modality-shared projectors are used to inject prompts into dual encoders. \textbf{Right-Top}: The low-rank adapter is decomposed into magnitude $\alpha_i$ and direction components, with historical directions frozen to preserve early geometric alignments, while the magnitudes remain trainable. \textbf{Right-Bottom}: The knowledge constancy loss $\mathcal{L}_{kcl}$ prevents feature drift and preserves pretrained semantic structure.}
    \label{fig:example}
\end{figure*}

\subsubsection{Learnable Embedding Space}
We first initialize a unified, learnable embedding space $E \in \mathbb{R}^{K \times d_r}$, where $K$ vectors are sampled from a zero-mean Gaussian distribution $\mathcal{N}(0, \sigma^2)$. This shared embedding is then transformed into modality-specific prompts for each layer through a projector. Specifically, prompts are inserted starting from a predefined layer $J$ ($1 \leq J \leq L$). For modality $m \in \{v, t\}$ at each layer $i \in \{J, \dots, L\}$, the prompt $P_i^m \in \mathbb{R}^{l \times d_m}$ is generated as:
\begin{equation}
    P_i^m = \text{Proj}_i^m(E),
    \label{eq:prompt_gen_func}
\end{equation}
where $\text{Proj}_i^m(\cdot)$ denotes a learnable projection function tailored for layer $i$ and modality $m$. The generated prompts $P_i^m$ are then concatenated with the original input tokens at the corresponding layer.

\subsubsection{Decoupled Low-Rank Expansion}
To efficiently model both the cross-layer semantic patterns and the layer-specific adaptations, we propose a parameter-efficient projection mechanism inspired by LoRA~\cite{hu2021lora}. The core idea is to decouple the projector's weight matrix into a shared component and a low-rank, layer-wise adapter. Typically, LoRA updates a frozen weight $W_0 \in \mathbb{R}^{d \times k}$ via a low-rank residual $BA$, where $B \in \mathbb{R}^{d \times r}$ and $A \in \mathbb{R}^{r \times k}$ ($r \ll \min(d, k)$). Extending this, for each projector associated with layer $i \in \{J, \dots, L\}$, the projector weight matrix is first decomposed as:
\begin{equation}
    W_{i}^{m} = W_{\text{shared}}^{m} + \Delta W_{i}^{m},
    \label{eq:decomposition_final_revised}
\end{equation}
where $W_{\text{shared}}^{m} \in \mathbb{R}^{d_{r} \times d_{m}}$ is a modality-specific shared component maintained across layers from $J$ to $L$ to capture fundamental semantic knowledge and alleviate redundancy. For notational brevity, we omit the modality superscript $m$ in the following. The layer-specific adapter $\Delta W_{i}$ is then parameterized via a low-rank decomposition, leading to the final form:
\begin{equation}
    W_{i} = W_{\text{shared}} + A_{i} B_{i},
    \label{eq:lora_decomposition_revised}
\end{equation}
where $A_{i} \in \mathbb{R}^{d_{r} \times r}$ and $B_{i} \in \mathbb{R}^{r \times d_{m}}$ are trainable low-rank matrices.
The adoption of a single $W_{\text{shared}}$ coupled with layer-wise low-rank adapters enables EvoPrompt to retain expressive power with significantly fewer parameters. Consequently, the parameter complexity is lowered from $\mathcal{O}((L-J+1) \cdot d_r d_m)$ to $\mathcal{O}(d_r d_m + (L-J+1) \cdot r(d_r + d_m))$. This formulation naturally enforces structural alignment through the shared base and strengthens generalization by cleanly separating common knowledge from layer-specific adjustments.

\subsection{Evolutionary Trajectory-Aware Learning Strategy}
\label{sec:decoupling}

While the MPP architecture establishes structural bridges across layers, optimizing prompts throughout the training process remains challenging due to the risk of catastrophic forgetting. We observe that prompts, which serve as general contextual anchors in early training stages, tend to converge toward task-specific patterns in later epochs, potentially overwriting previously acquired generalizable knowledge. To mitigate this, inspired by progressive learning methods~\cite{gong2019efficient, yano2025efficient}, we introduce an evolutionary trajectory-aware learning strategy. This approach explicitly decouples and modulates parameter effects learned at different phases, conceptualizing adaptation as a progressive accumulation of knowledge.

\subsubsection{Incremental Magnitude-Direction Decoupling}
Building on insights from weight decomposition analysis~\cite{wu2025sd}, we factorize the layer-wise low-rank update $\Delta W_i^t$ at training epoch $t$ into a learnable magnitude coefficient $\alpha_i^t$ and a normalized directional matrix:
\begin{equation}
    \Delta W_i^t = \alpha_i^t \cdot \frac{\mathbf{A}_i^t \mathbf{B}_i^t}{\|\mathbf{A}_i^t \mathbf{B}_i^t\|_F} = \alpha_i^t \cdot \overline{\mathbf{A}_i^t \mathbf{B}_i^t},
    \label{eq:decoupling}
\end{equation}
where $\|\cdot\|_F$ denotes the Frobenius norm. This explicit decoupling grants independent control over the adaptation strength ($\alpha_i^t$) and its direction ($\overline{\mathbf{A}_i^t \mathbf{B}_i^t}$). To promote stable progressive learning, we frame the training process as the accumulation of directional knowledge. Prior work suggests the directional component is more critical than its magnitude in low-rank adaptation~\cite{liu2024dora, qiu2023controlling}. Accordingly, we extend Eq.~\eqref{eq:lora_decomposition_revised} to compute the adapter weight for layer $i$ at epoch $T$ as a historical sum:
\begin{equation}
    W_{i}^{T} = W_{\text{shared}} + \sum_{t=1}^{T-1} \textcolor{orange}{\alpha_{i}^t} \overline{\mathbf{A}_{i}^t \mathbf{B}_{i}^t} + \textcolor{orange}{\alpha_{i}^T} \textcolor{orange}{\overline{\mathbf{A}_{i}^T \mathbf{B}_{i}^T}},
    \label{eq:incremental_update}
\end{equation}
Here, $W_{\text{shared}}$ acts as a fixed, modality-specific foundation. During training at epoch $T$, we freeze all previously acquired directions $\{\overline{\mathbf{A}_i^t \mathbf{B}_i^t}\}_{t=1}^{T-1}$ to preserve their geometric structure. Only the magnitude coefficients $\{\alpha_i^t\}_{t=1}^T$ and the new direction $\overline{\mathbf{A}_i^T \mathbf{B}_i^T}$ remain trainable. This design enables the model to recalibrate the influence of past knowledge via the learnable $\alpha_i^t$ while progressively incorporating new directional adjustments, thereby adapting to the evolving loss landscape without catastrophically forgetting previously learned, robust features.

\subsubsection{Adaptive Rank Reduction}
To enhance continual adaptation stability and mitigate the risk of overfitting during later evolutionary stages, we introduce an empirical rank-reduction mechanism. Denoting the total number of training epochs as $N_e$, this strategy modulates the capacity of the learnable matrices $\mathbf{A}_{i}^t \in \mathbb{R}^{d_r \times r^t}$ and $\mathbf{B}_{i}^t \in \mathbb{R}^{r^t \times d_m}$ by adjusting the rank $r^t$ in a controlled, stepwise manner:
\begin{equation}
    r^1 = r^2 = \dots > r^{\mu} = r^{\mu+1} = \dots > r^{\nu} = r^{\nu+1} = \dots = r^{N_e},
    \label{eq:rank_reduction}
\end{equation}
where $\mu$ and $\nu$ ($1 < \mu < \nu \leq N_e$) represent predefined epoch indices at which the rank is reduced. By assigning lower-rank weights to later epochs based on their diminishing marginal contributions, this strategy imposes a structural regularization that stabilizes the optimization landscape. Consequently, it significantly reduces cumulative computational and memory overhead while maintaining the model's generalization ability.

\subsection{Overall Training Objective}
To preserve the rich semantic knowledge embedded in the pre-trained CLIP model, we introduce a knowledge constancy loss on both modalities. Let $f^v$ and $f^t$ be the prompted features for a single sample, and let $f^v_0$ and $f^t_0$ denote the corresponding features extracted by the original, frozen CLIP encoders (without prompts). The constancy loss is formulated as:
\begin{equation}
\mathcal{L}_{kcl} = \frac{1}{2} \left[ \left(1 - \frac{{f^v}^\top f^v_0}{\|f^v\| \|f^v_0\|}\right) + \left(1 - \frac{{f^t}^\top f^t_0}{\|f^t\| \|f^t_0\|}\right) \right].
\label{eq:kcl_loss}
\end{equation}
This term ensures that the learned prompts do not cause the feature representations to deviate excessively from the well-structured original CLIP feature distribution, thereby maintaining its strong zero-shot generalization capability.

Combining the standard contrastive alignment loss $\mathcal{L}_{InfoNCE}$ and the knowledge constancy loss, our complete training objective is:
\begin{equation}
\mathcal{L}_{total} = \mathcal{L}_{InfoNCE} + \eta \mathcal{L}_{kcl},
\label{eq:total_loss}
\end{equation}
where $\eta$ is a balancing hyperparameter. This composite loss guides the model to achieve strong cross-modal instance alignment while preserving the pre-trained feature geometry. A complete algorithmic flowchart illustrating our training strategy is summarized in Algorithm \ref{alg:evoprompt}.

\begin{algorithm}[ht]
\caption{Evolutionary Trajectory-Aware Learning}
\label{alg:evoprompt}

\KwIn{Pre-trained encoders $f_0^v, f_0^t$; embedding $E \in \mathbb{R}^{K \times d_r}$;
projector weights $W_{\text{shared}}^m$; rank indices $\mu, \nu$; hyperparameter $\eta$.}
\KwOut{$E,\ W_{\text{shared}}^m,\ \{\alpha_i^k\},\ \mathcal{W}$.}

\BlankLine

$\mathcal{W} \leftarrow \emptyset$

\BlankLine

\For{each training epoch $T$}{
    \If{$T = \mu$ \textbf{or} $T = \nu$}{
        Reduce rank $r^T$
    }

    Freeze all directional matrices in $\mathcal{W}$

    \ForEach{layer $i \in \{J, \dots, L\}$}{
        $\Delta W_i^T = \sum_{t=1}^{T-1} \alpha_i^t \overline{\mathbf{A}_i^t \mathbf{B}_i^t}
        + \alpha_i^T \overline{\mathbf{A}_i^T \mathbf{B}_i^T}$

        $P_i^m = E \cdot (W_{\text{shared}}^m + \Delta W_i^T)$
    }

    Inject $\{P_i^m\}_{i=J}^L$; extract $\mathcal{F}^v, \mathcal{F}^t$ and frozen $f_0^v, f_0^t$\;

    $\mathcal{L}_{total} = \mathcal{L}_{InfoNCE} + \eta\,\mathcal{L}_{kcl}$

    Update $E,\ W_{\text{shared}}^m,\ \{\alpha_i^T\},\ \mathbf{A}_i^T,\ \mathbf{B}_i^T$\;

    \ForEach{layer $i \in \{J, \dots, L\}$}{
        $\mathcal{W} \leftarrow \mathcal{W} \cup \{\overline{\mathbf{A}_i^T \mathbf{B}_i^T}\}$
    }
}

\BlankLine

\Return{$E,\ W_{\text{shared}}^m,\ \{\alpha_i^k\},\ \mathcal{W}$}
\end{algorithm}

\section{Experiment}
\label{sec:experiments}

We evaluate EvoPrompt under four standard experimental settings: base-to-novel generalization, cross-dataset transfer, domain generalization, and few-shot learning. Unless otherwise noted, we strictly follow the evaluation protocols established in prior work~\cite{zhou2022learning, zhou2022conditional}.

\subsection{Tasks and Datasets}
\label{sec:tasks_datasets}

\subsubsection{Base-to-Novel Generalization}
To evaluate the trade-off between task-specific adaptation and zero-shot capability preservation, we split the categories of each dataset equally into a base set for training and a novel set for evaluation. Models are trained exclusively on the base classes and tested on both base and novel classes. This experiment is conducted across 11 standard image classification benchmarks: ImageNet~\cite{deng2009imagenet}, Caltech101~\cite{fei2004learning}, OxfordPets~\cite{parkhi2012cats}, StanfordCars~\cite{krause20133d}, Flowers102~\cite{nilsback2008automated}, Food101~\cite{bossard2014food}, FGVCAircraft~\cite{maji2013fine}, SUN397~\cite{xiao2010sun}, UCF101~\cite{soomro2012ucf101}, DTD~\cite{cimpoi2014describing}, and EuroSAT~\cite{helber2019eurosat}.
\subsubsection{Cross-Dataset Transfer}
Following CoCoOp~\cite{zhou2022conditional}, we assess out-of-distribution generalization by training a 16-shot model on ImageNet (covering all 1,000 classes) and then evaluating the frozen model directly on the other 10 datasets without any further fine-tuning.
\subsubsection{Domain Generalization}
We measure robustness to distribution shifts by evaluating the same ImageNet-trained model on four challenging ImageNet variants: ImageNet-V2~\cite{recht2019do}, ImageNet-Sketch~\cite{wang2019learning}, ImageNet-A~\cite{hendrycks2021natural}, and ImageNet-R~\cite{hendrycks2021many}. This evaluates the model's ability to maintain performance under domain shift.
\subsubsection{Few-Shot Learning}
To evaluate sample efficiency, we train models with varying numbers of labeled examples (1, 2, 4, 8, and 16 shots per category) and test on the full test sets of each benchmark. This setting probes the model's ability to learn effectively from extremely limited supervision and reveals whether it acquires both task-specific discriminative patterns and task-agnostic knowledge.

\begin{table*}[ht!]
\centering

\caption{Comparison with state-of-the-art methods on base-to-novel generalization across 11 datasets. EvoPrompt demonstrates strong generalization results over existing methods. The best results are in bold and the second-best results are underlined.}
\label{tab:base2novel_comparison}
\begin{tabular}{r|ccc|ccc|ccc|ccc}
\toprule
\multirow{2}{*}{Method} & \multicolumn{3}{c|}{Average} & \multicolumn{3}{c|}{ImageNet} & \multicolumn{3}{c|}{Caltech101} & \multicolumn{3}{c}{OxfordPets} \\
& Base & Novel & HM & Base & Novel & HM & Base & Novel & HM & Base & Novel & HM \\
\midrule
CLIP & 69.34 & 74.22 & 71.70 & 72.43 & 68.14 & 70.22 & 96.84 & 94.00 & 95.40 & 91.17 & 97.26 & 94.12 \\
CoOp & 82.69 & 63.22 & 71.66 & 76.47 & 67.88 & 71.92 & 98.00 & 89.81 & 93.73 & 93.67 & 95.29 & 94.47 \\
CoCoOp & 80.47 & 71.69 & 75.83 & 75.98 & 70.43 & 73.10 & 97.96 & 93.81 & 95.84 & 95.20 & 97.69 & 96.43 \\
ProDA & 81.56 & 72.30 & 76.65 & 75.40 & 70.23 & 72.72 & 98.27 & 93.23 & 95.68 & 95.43 & 97.83 & 96.62 \\
KgCoOp & 80.73 & 73.60 & 77.00 & 75.83 & 69.96 & 72.78 & 97.72 & 94.39 & 96.03 & 94.65 & 97.76 & 96.18 \\
MaPLe & 82.28 & 75.14 & 78.55 & 76.66 & 70.54 & 73.47 & 97.74 & 94.36 & 96.02 & 95.43 & 97.76 & 96.58 \\
PromptSRC & 84.26 & 76.10 & \underline{79.97} & \underline{77.60} & 70.73 & 74.01 & 98.10 & 94.03 & 96.02 & 95.33 & 97.30 & 96.30 \\
ProVP & \underline{85.20} & 73.22 & 78.76 & 75.82 & 69.21 & 72.36 & \underline{98.92} & 94.21 & \underline{96.51} & \underline{95.87} & 97.65 & \underline{96.75} \\
MetaPrompt & 83.65 & 75.48 & 79.09 & 77.52 & 70.83 & 74.02 & 98.13 & 94.58 & 96.32 & 95.53 & 97.00 & 96.26 \\
TCP & 84.13 & 75.36 & 79.51 & 77.27 & 69.87 & 73.38 & 98.23 & \underline{94.67} & 96.42 & 94.67 & 97.20 & 95.92 \\
MMA & 83.20 & \underline{76.80} & 79.87 & 77.31 & \underline{71.00} & \underline{74.02} & 98.40 & 94.00 & 96.15 & 95.40 & \underline{98.07} & 96.72 \\
\midrule
\rowcolor{gray!20} EvoPrompt & \textbf{85.90} & \textbf{78.35} & \textbf{81.95} & \textbf{77.98} & \textbf{72.10} & \textbf{74.92} & \textbf{99.11} & \textbf{95.30} & \textbf{97.17} & \textbf{96.13} & \textbf{98.40} & \textbf{97.25} \\

\midrule\midrule

\multirow{2}{*}{Method} & \multicolumn{3}{c|}{StanfordCars} & \multicolumn{3}{c|}{Flowers102} & \multicolumn{3}{c|}{Food101} & \multicolumn{3}{c}{FGVCAircraft} \\
& Base & Novel & HM & Base & Novel & HM & Base & Novel & HM & Base & Novel & HM \\
\midrule
CLIP & 63.37 & 74.89 & 68.65 & 72.08 & \underline{77.80} & 74.83 & 90.10 & 91.22 & 90.66 & 27.19 & 36.29 & 31.09 \\
CoOp & 78.12 & 60.40 & 68.13 & 97.60 & 59.67 & 74.06 & 88.33 & 82.26 & 85.19 & 40.44 & 22.30 & 28.75 \\
CoCoOp & 70.49 & 73.59 & 72.01 & 94.87 & 71.75 & 81.71 & 90.70 & 91.29 & 90.99 & 33.41 & 23.71 & 27.74 \\
ProDA & 74.70 & 71.20 & 72.91 & 97.70 & 68.68 & 80.66 & 90.30 & 88.57 & 89.43 & 36.90 & 34.13 & 35.46 \\
KgCoOp & 71.76 & \underline{75.04} & 73.36 & 95.00 & 74.73 & 83.65 & 90.50 & 91.70 & 91.09 & 36.21 & 33.55 & 34.83 \\
MaPLe & 72.94 & 74.00 & 73.47 & 95.92 & 72.46 & 82.56 & 90.71 & \underline{92.05} & \underline{91.38} & 37.44 & 35.61 & 36.50 \\
PromptSRC & 78.27 & 74.97 & 76.58 & 98.07 & 76.50 & \underline{85.95} & 90.67 & 91.53 & 91.10 & 42.73 & \underline{37.87} & \underline{40.15} \\
ProVP & 80.43 & 67.96 & 73.67 & \underline{98.42} & 72.06 & 83.20 & 90.32 & 90.91 & 90.61 & \underline{47.08} & 29.87 & 36.55 \\
MetaPrompt & 76.34 & 75.01 & 75.48 & 97.66 & 74.49 & 84.52 & \underline{90.74} & 91.85 & 91.29 & 40.14 & 36.51 & 38.24 \\
TCP & \underline{80.80} & 74.13 & \underline{77.32} & 97.73 & 75.57 & 85.23 & 90.57 & 91.37 & 90.97 & 41.97 & 34.43 & 37.83 \\
MMA & 78.50 & 73.10 & 75.70 & 97.77 & 75.93 & 85.48 & 90.13 & 91.30 & 90.71 & 40.57 & 36.33 & 38.33 \\
\midrule
\rowcolor{gray!20} EvoPrompt & \textbf{81.15} & \textbf{75.90} & \textbf{78.44} & \textbf{98.70} & \textbf{78.33} & \textbf{87.34} & \textbf{91.10} & \textbf{92.78} & \textbf{91.93} & \textbf{47.42} & \textbf{39.54} & \textbf{43.12} \\

\midrule\midrule

\multirow{2}{*}{Method} & \multicolumn{3}{c|}{SUN397} & \multicolumn{3}{c|}{DTD} & \multicolumn{3}{c|}{EuroSAT} & \multicolumn{3}{c}{UCF101} \\
& Base & Novel & HM & Base & Novel & HM & Base & Novel & HM & Base & Novel & HM \\
\midrule
CLIP & 69.36 & 75.35 & \underline{72.23} & 53.24 & 59.90 & 56.37 & 56.48 & 64.05 & 60.03 & 70.53 & 77.50 & 73.85 \\
CoOp & 80.60 & 65.89 & 72.51 & 79.44 & 41.18 & 54.24 & 92.19 & 54.74 & 68.69 & 84.69 & 56.05 & 67.46 \\
CoCoOp & 79.74 & 76.86 & 78.27 & 77.01 & 56.00 & 64.85 & 87.49 & 60.04 & 71.21 & 82.33 & 73.45 & 77.64 \\
ProDA & 78.67 & 76.93 & 77.79 & 80.67 & 56.48 & 66.44 & 83.90 & 66.00 & 73.88 & 85.23 & 71.97 & 78.04 \\
KgCoOp & 80.29 & 76.53 & 78.36 & 77.55 & 54.99 & 64.35 & 85.64 & 64.34 & 73.48 & 82.89 & 76.67 & 79.65 \\
MaPLe & 80.82 & 78.70 & 79.75 & 80.36 & 59.18 & 68.16 & 94.07 & 73.23 & 82.35 & 83.00 & 78.66 & 80.77 \\
PromptSRC & \underline{82.67} & 78.47 & 80.52 & 83.37 & 62.97 & 71.75 & 92.90 & 73.90 & 82.32 & 87.10 & 78.80 & 82.74 \\
ProVP & 80.67 & 76.11 & 78.32 & \underline{83.95} & 59.06 & 69.34 & \underline{97.12} & 72.91 & 83.29 & \textbf{88.56} & 75.55 & 81.54 \\
MetaPrompt & 82.26 & \underline{79.04} & \underline{80.62} & 83.10 & 58.05 & 68.35 & 93.53 & 75.21 & 83.38 & 85.33 & 77.72 & 81.35 \\
TCP & 82.63 & 78.20 & 80.35 & 82.77 & 58.07 & 68.25 & 91.63 & 74.73 & 82.32 & 87.13 & \underline{80.77} & \underline{83.83} \\
MMA & 82.27 & 78.57 & 80.38 & 83.20 & \underline{65.63} & \underline{73.38} & 85.46 & \underline{82.34} & \underline{83.87} & 86.23 & 80.03 & 82.20 \\
\midrule
\rowcolor{gray!20} EvoPrompt & \textbf{83.31} & \textbf{79.50} & \textbf{81.36} & \textbf{84.20} & \textbf{65.92} & \textbf{73.95} & \textbf{97.43} & \textbf{82.79} & \textbf{89.52} & \underline{88.34} & \textbf{81.30} & \textbf{84.67} \\
\bottomrule
\end{tabular}
\vspace{-1em}
\end{table*}

\subsection{Implementation Details}
We adopt a pre-trained CLIP with a ViT-B/16~\cite{dosovitskiy2020image} backbone as our foundation model. Unless investigating variable-shot performance, we sample 16 shots per class following prior work~\cite{zhou2022learning, khattak2023maple, yang2024mma, zhou2022conditional, yao2023visual, zhu2023prompt, yao2024tcp}. Zero-shot classifier weights are generated using standard prompt templates~\cite{radford2021learning, zhou2022learning, zhang2022tip}. Both the visual encoder $F$ and the text encoder $G$ remain fully frozen. For the learnable embedding space $E$, we set the number of vectors to $K=5$ and the shared representation dimension to $d_r=512$. Prompts with token length $l=5$ are inserted from layer $J=6$ to the final layer $L=12$. For the rank reduction mechanism, we set $\mu{=}4$ and $\nu{=}7$ with $N_e{=}10$, reducing the rank as $r^1{=}8 \to r^{\mu}{=}4 \to r^{\nu}{=}2$. The more detailed configurations and analysis of these parameters, along with the $r^t$, $\mu$ and $\nu$ in rank reduction mechanism, are provided in the supplementary materials. All experiments are conducted on a single NVIDIA A800 GPU, and we report the average top-1 accuracy over three random seeds.

\subsection{Base-to-Novel Generalization}
In this experiment, we evaluate the performance of EvoPrompt against several representative benchmarks, including the zero-shot CLIP baseline and various state-of-the-art prompt learning techniques such as CoOp \cite{zhou2022learning}, CoCoOp \cite{zhou2022conditional}, ProDA \cite{proda}, KgCoOp \cite{yao2023visual}, MaPLe \cite{khattak2023maple}, PromptSRC \cite{khattak2023self}, ProVP \cite{xu2025progressive}, MetaPrompt \cite{zhao2024learning}, TCP \cite{yao2024tcp}, and the adapter-based method MMA \cite{yang2024mma}. To ensure a fair comparison, we exclude methods that rely on large language models for external prompt priors or those utilizing full unlabeled datasets for distillation.

Table \ref{tab:base2novel_comparison} presents a comprehensive comparison highlighting the superior average performance of EvoPrompt across 11 datasets. In particular, our method surpasses the leading baselines by 1.55\% on Novel classes and 1.82\% in HM. Additionally, the results highlight an advanced transfer learning capability, evidenced by substantial gains in base accuracy alongside sustained strong generalizability.

\begin{table*}[t!]
\centering
\caption{Comparison of Ours with previous state-of-the-art methods on cross-dataset evaluation across 10 datasets.}
\label{tab:cross_dataset}
\resizebox{\linewidth}{!}{
\begin{tabular}{rc|cccccccccc | c} 
\toprule
\multirow{2}{*}{\textbf{Method}} & \textbf{Source} & \multicolumn{11}{c}{\textbf{Target}} \\
\cmidrule(r){2-2} \cmidrule(l){3-13} 
& \rotatebox{60}{ImageNet} & \rotatebox{60}{Caltech101} & \rotatebox{60}{OxfordPets} & \rotatebox{60}{StanfordCar} & \rotatebox{60}{Flowers102} & \rotatebox{60}{Food101} & \rotatebox{60}{Aircraft} & \rotatebox{60}{SUN397} & \rotatebox{60}{DTD} & \rotatebox{60}{EuroSAT} & \rotatebox{60}{UCF101} & \rotatebox{60}{\textit{Average}} \\
\midrule
CoOp & \underline{71.51} & 93.70 & 89.14 & 64.51 & 68.71 & 85.30 & 18.47 & 64.15 & 41.92 & 46.39 & 66.55 & 63.88 \\
CoCoOp & 71.02 & \underline{94.43} & 90.14 & 65.32 & 71.88 & 86.06 & 22.94 & 67.36 & 45.73 & 45.37 & 68.21 & 65.74 \\
MaPLe & 70.72 & 93.53 & 90.49 & 65.57 & \underline{72.23} & 86.20 & 24.74 & 67.01 & 46.49 & 48.06 & 68.69 & 66.30 \\
PromptSRC & 71.27 & 93.60 & 90.25 & 65.70 & 70.25 & 86.15 & 23.90 & 67.10 & \underline{46.87} & 45.50 & \underline{68.75} & 65.81 \\
TCP & 71.40 & 93.97 & \textbf{91.25} & 64.69 & 71.21 & \underline{86.69} & 23.45 & 67.15 & 44.35 & \textbf{51.45} & 68.73 & 66.29 \\
MMA & 71.00 & 93.80 & 90.30 & \underline{66.13} & 72.07 & 86.12 & \underline{25.33} & \underline{68.17} & 46.57 & 49.24 & 68.32 & \underline{66.61} \\
\midrule
\rowcolor{gray!15} \textbf{EvoPrompt} & \textbf{72.10} & \textbf{95.12} & \underline{90.55} & \textbf{66.25} & \textbf{72.78} & \textbf{87.65} & \textbf{25.50} & \textbf{68.32} & \textbf{47.15} & \underline{51.20} & \textbf{69.55} & \textbf{67.41} \\
\bottomrule
\end{tabular}
}
\end{table*}
\subsection{Cross-Dataset Evaluation}
Tab. \ref{tab:cross_dataset} presents the cross-dataset evaluation results, where all models are trained on ImageNet and directly evaluated across 10 diverse target datasets. Notably, EvoPrompt achieves the highest average target accuracy of 67.41\%, outperforming leading baselines such as MMA (66.61\%) and MaPLe (66.30\%). Concurrently, it attains the top source accuracy on ImageNet (72.10\%). This dual superiority suggests that the shared representation space of MPP, coupled with our evolutionary learning strategy, yields highly transferable prompts that generalize better than MaPLe's independently parameterized per-layer design. Overall, EvoPrompt achieves the best balance between source-domain performance and cross-dataset transferability among all compared methods.
\subsection{Domain Generalization}
Tab.~\ref{tab:domain_gen} assesses model robustness under natural distribution shifts using four challenging ImageNet variant datasets. EvoPrompt achieves the best average accuracy across all domains, demonstrating that it not only enhances in-distribution classification but also more effectively preserves CLIP's inherent out-of-distribution generalization capabilities compared to existing adaptation methods.

\begin{table}
    \centering
    \caption{Comparison of EvoPrompt with previous methods on domain generalization across 4 datasets.}
    \label{tab:domain_gen}
    \resizebox{\linewidth}{!}{%
    \footnotesize
    \begin{tabular}{rc|cccc}
    \toprule
    ~ & Source & \multicolumn{4}{c}{Target} \\
    \cmidrule{2-6}
    & ImNet & -V2 & -S & -A & -R \\
    \midrule
    CLIP & 66.73 & 60.83 & 46.15 & 47.77 & 73.96 \\
    CoOp & \underline{71.51} & 64.20 & 47.99 & 49.71 & 75.21 \\
    CoCoOp & 71.02 & 64.07 & 48.75 & 50.63 & 76.18 \\
    MaPLe & 70.72 & 64.07 & 49.15 & 50.90 & 76.98 \\
    PromptSRC & 71.27 & \underline{64.35} & \underline{49.55} & 50.90 & \underline{77.80} \\
    MMA & 71.00 & 64.33 & 49.13 & \underline{51.12} & 77.32 \\
    \midrule
    \rowcolor{gray!20} EvoPrompt & \textbf{72.10} & \textbf{64.40} & \textbf{49.72} & \textbf{51.30} & \textbf{77.90} \\
    \bottomrule
    \end{tabular}
    }
\end{table}

\begin{table}[t]
  \centering
  \caption{Cumulative component ablation on ImageNet. 
           Each row adds one design choice to the previous 
           configuration. \textbf{Bold} denotes the best result.}
  \label{tab:ablation_cumulative}
  \setlength{\tabcolsep}{8pt}
  \renewcommand{\arraystretch}{1.15}
  \begin{tabular}{lccc}
    \toprule
    \textbf{Configuration} & \textbf{Base} & \textbf{Novel} & \textbf{HM} \\
    \midrule
    Baseline       & 74.21 & 67.83 & 70.88 \\
    + MPP                         & 75.34 & 69.12 & 72.09 \\
    + $W_{\text{shared}}$ + AB    & 75.89 & 70.01 & 72.83 \\
    + E.T. w/o Rank Reduction    & 76.73 & 70.50 & 73.48 \\
    + E.T.  & 76.45 & 70.98 & 73.61 \\
    \rowcolor{gray!15}
    + $\mathcal{L}_{kcl}$ \textit{(full)}
                                        & \textbf{77.98} & \textbf{72.10} & \textbf{74.92} \\
    \bottomrule
  \end{tabular}
\end{table}

\begin{figure*}[t!] 
    \centering 
    \includegraphics[width=1.0\textwidth]{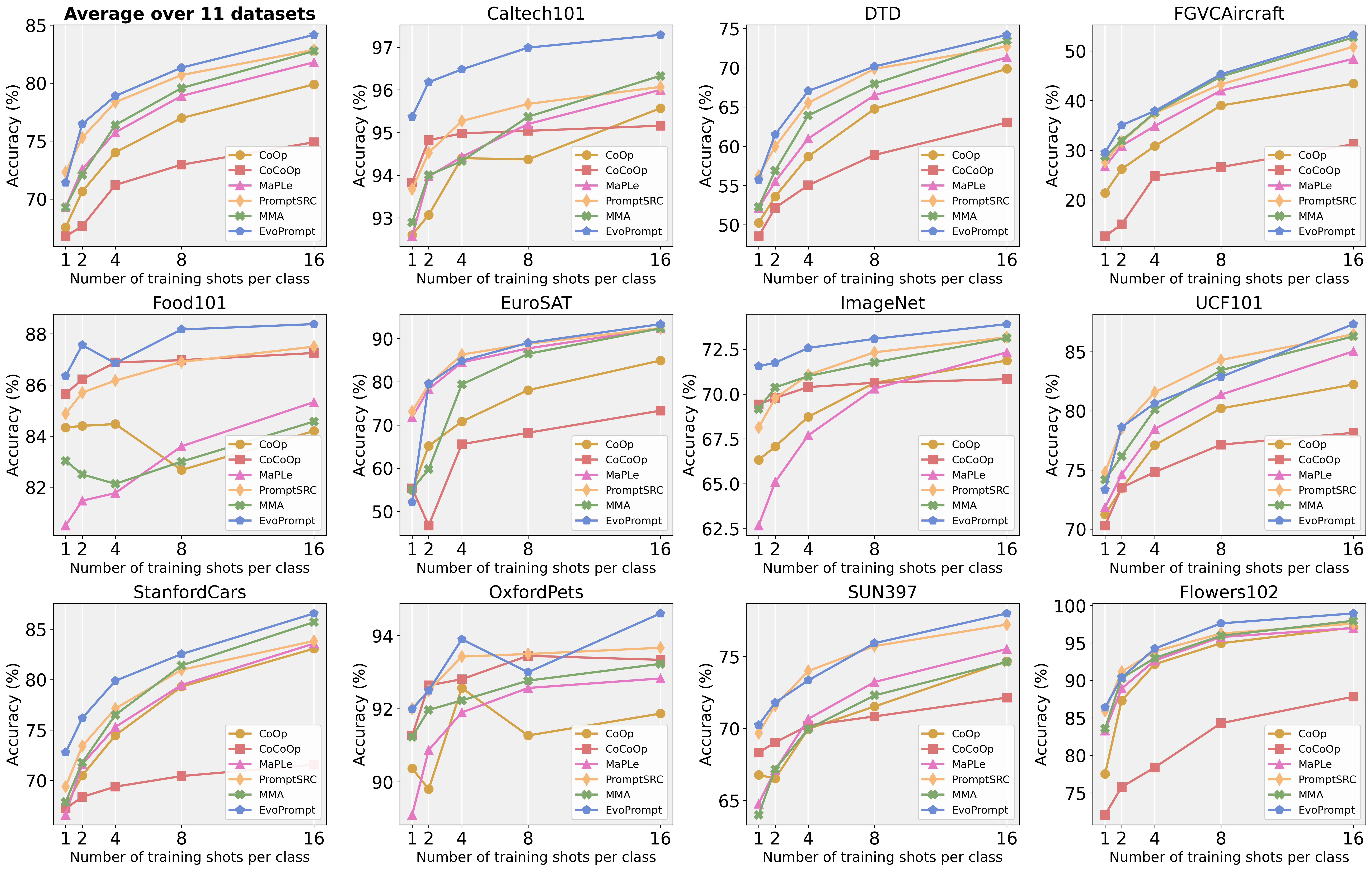} 
    \caption{EvoPrompt performance comparison in few-shot image recognition setting.} 
    \label{fig:few_shot} 
\end{figure*}

\subsection{Few-Shot Learning}
Few-shot classification results are presented in Fig.~\ref{fig:few_shot}. EvoPrompt demonstrates solid and competitive performance across the evaluated shot settings. While performance is broadly comparable in the most data-scarce regimes, the advantage of EvoPrompt becomes more pronounced as the number of training examples increases. This scaling behavior indicates that the framework effectively leverages additional supervisory signals to learn increasingly transferable representations.

\subsection{Ablation Study}
\label{sec:ablation}
\subsubsection{Component Analysis}
Tab.~\ref{tab:ablation_cumulative} presents a cumulative ablation study on ImageNet, where each row incrementally integrates a specific design choice to demonstrate the consistent and complementary contributions of individual modules. Starting from the baseline configuration with isolated, independent prompts (70.88\% HM), introducing MPP yields the largest single-step improvement (+1.21\% HM), confirming that cross-layer and cross-modal information sharing serves as a critical structural foundation. The subsequent addition of the shared weight matrix $W_{\text{shared}}$ and the low-rank adapter (AB) contributes an additional +0.74\% HM, validating that our decoupled low-rank parameterization is both parameter-efficient and expressive. Furthermore, incorporating the evolutionary training strategy (E.T.) yields a further +0.78\% HM gain, effectively mitigating catastrophic forgetting while preserving robust generalization capabilities. Notably, the adaptive rank reduction mechanism itself contributes a measurable +0.13\% HM gain (73.48\% $\to$ 73.61\%); removing it causes the model to overfit the base distribution (76.73\% Base) at the expense of novel-class generalization (70.50\% Novel), confirming its role as an effective late-stage regularizer. Finally, integrating the knowledge constancy loss $\mathcal{L}_{kcl}$ delivers a substantial +1.31\% HM improvement, demonstrating that constraining feature drift provides significant complementary benefits on top of all preceding components. In its full configuration, EvoPrompt achieves 74.92\% HM, representing a total cumulative advancement of +4.04\% over the baseline.

\begin{table}[t]
    \centering
    \caption{Sensitivity analysis of loss weights $\eta$ for 
             $\mathcal{L}_{kcl}$
             on ImageNet. \textbf{Bold} denotes the best result.}
    \label{tab:ablation_weights}
    \setlength{\tabcolsep}{8pt}
    \renewcommand{\arraystretch}{1.1}
    \begin{tabular}{llccc}
        \toprule
        \textbf{Loss} & \textbf{Value} & \textbf{Base} & \textbf{Novel} & \textbf{HM} \\
        \midrule
        \multirow{4}{*}{$\eta$}
            & 0.2 & 77.05 & 71.48 & 74.17 \\
            & 0.5 & \textbf{77.98} & \textbf{72.10} & \textbf{74.92} \\
            & 1.0 & 76.83 & 71.73 & 74.18 \\
            & 2.0 & 76.45 & 71.30 & 73.80 \\
        \bottomrule
    \end{tabular}
\end{table}

\begin{table}[t]
    \centering
    \caption{Comparison of training efficiency on ImageNet. 
             Trainable parameters (M), training time (ms/image), 
             and FPS (batch size=100) are reported.}
    \label{tab:ablation_efficiency}
    \setlength{\tabcolsep}{8pt}
    \renewcommand{\arraystretch}{1.1}
    \begin{tabular}{lccc}
        \toprule
        \textbf{Method} & \textbf{Params (M)} & \textbf{Time (ms)} & \textbf{FPS} \\
        \midrule
        MaPLe  & 3.555 & 39.5 & 1757.6 \\
        PSRC   & 0.046 & 40.0 & 1764.2 \\
        ProVP  & 0.147 &  6.4 &  928.9 \\
        MetaP  & 0.031 & 30.7 &  659.8 \\
        TCP    & 0.332 &  5.3 &  950.6 \\
        MMA    & 0.675 &  2.2 &  688.5 \\
        \midrule
        \rowcolor{gray!15}
        EvoPrompt & 0.764 & 4.5 & 1282.1 \\
        \bottomrule
    \end{tabular}
\end{table}

\subsubsection{Loss Weight Sensitivity}
We evaluate the sensitivity of the hyper-parameter $\eta$ (for $\mathcal{L}_{kcl}$) on ImageNet, as shown in Tab.~\ref{tab:ablation_weights}. The model achieves the optimal trade-off at $\eta=0.5$, reaching 74.92\% HM. Deviating from these values causes a performance drop, particularly in the harmonic mean, suggesting that a balanced weighting of the proposed losses is crucial for robust cross-class generalization.
\begin{figure*}[t!]
    \centering
    
    \includegraphics[width=1.0\textwidth]{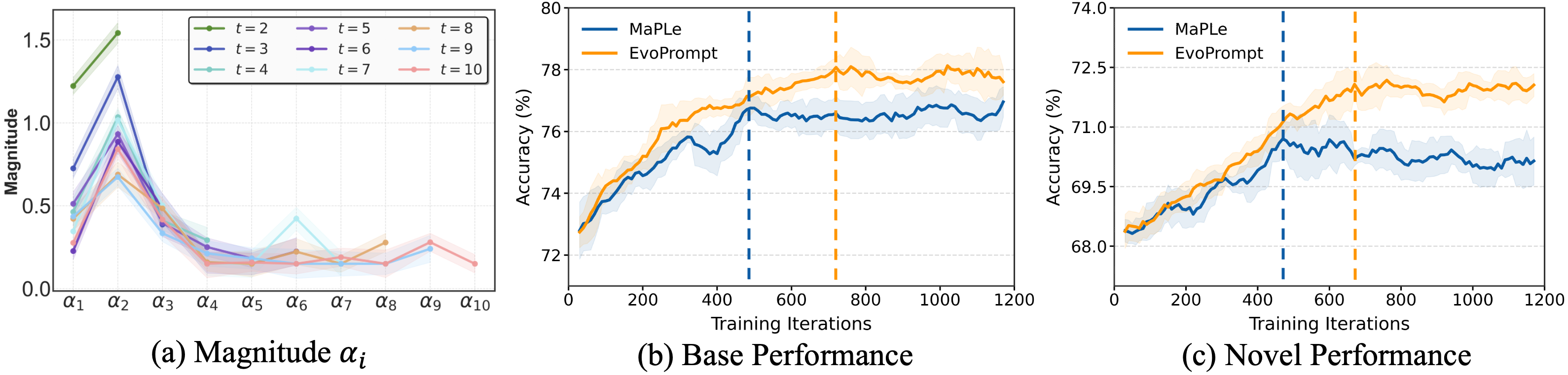}
    \caption{Analysis of training dynamics and performance. (a) The evolution of learnable magnitudes $\alpha_i$. (b, c) Performance comparison between MaPLe and EvoPrompt, where vertical dashed lines indicate training breakpoints.}
    \label{fig:combined_analysis}
\end{figure*}

\subsection{Further Analysis}
\subsubsection{Computational Efficiency}
As shown in Tab.~\ref{tab:ablation_efficiency}, all methods are benchmarked on a single NVIDIA A800 GPU under a unified protocol: the reported training time covers both forward and backward passes at batch size 100, while FPS measures inference throughput with a single forward pass at the same batch size. Under this protocol, EvoPrompt requires only 0.764M trainable parameters when trained for 10 epochs, which is comparable to or fewer than most efficient prior methods. Meanwhile, it attains a fast inference speed of 1282.1 FPS and requires only 4.5ms of training time per image. This efficiency stems from our lightweight design. The decoupled MPP structure cuts parameters by 4.6× compared to MaPLe. By freezing historical directional updates and optimizing only their magnitude coefficients, the expansion of learnable parameters throughout training remains minimal. Additionally, our adaptive rank reduction mechanism progressively decreases the rank of low-rank adapters in later epochs, naturally limiting parameter growth. Together, these strategies ensure that EvoPrompt maintains a lightweight and stable parameter footprint while delivering scalable adaptation performance.

\subsubsection{Evolution of Learned Magnitudes}
Analysis of the learned magnitude coefficients $\alpha$ on ImageNet reveals a distinct and stable evolutionary pattern across 10 training epochs, as shown in Fig.~\ref{fig:combined_analysis} (a). The values do not peak at the initial epoch $\alpha_1$, which can be attributed to the inherent instability at the start of training as the model begins to explore and adapt to the prompt space. Instead, they rise rapidly to a maximum at $\alpha_2$. This is followed by a gradual decline in later epochs. The pattern suggests a hierarchical importance: the prompt representation quickly consolidates core features using directions established early in training around $\alpha_2$, while later epochs contribute directions with diminishing magnitudes, primarily serving for fine-grained adjustment without drastically altering the established semantic space.

\subsubsection{Overfitting Phenomenon}
We analyze the training dynamics of MaPLe and EvoPrompt on the ImageNet, as illustrated in Fig.~\ref{fig:combined_analysis} (b,c). A critical “breakpoint” signifies a phase transition. Before this point, both methods learn transferable features, evidenced by joint performance gains on base and novel classes. After the breakpoint, MaPLe begins to over-specialize on the base training data, leading to unrecoverable performance degradation on novel classes despite improving base accuracy. In contrast, EvoPrompt maintains a stable and robust performance on novel classes after its breakpoint, effectively mitigating the overfitting issue. This demonstrates the superior generalization capability of EvoPrompt, achieved by optimizing the prompt evolution process.

\section{Conclusion}

In this work, we introduced EvoPrompt, a novel prompt tuning framework for few-shot adaptation of large vision-language models. At its core, a trajectory-aware training strategy decouples prompt evolution into frozen semantic directions and adaptive magnitudes, structurally grounded by MPP and complemented by a knowledge constancy loss that preserves pre-trained semantics. Experiments across diverse vision-language benchmarks confirm that governing the evolutionary trajectory of the prompts is the essence of effective and generalizable VLM adaptation.

\bibliographystyle{splncs04}
\bibliography{main}
\end{document}